%% file: main.tex
\documentclass[runningheads]{llncs}
\usepackage{graphicx}
\usepackage{comment}
\usepackage{makecell}
\usepackage{tabu}
\usepackage{amsmath,amssymb} 
\usepackage{color}
\usepackage{bbm}
\usepackage{booktabs}
\usepackage{subcaption}

\usepackage[pagebackref=true,breaklinks=true,letterpaper=true,colorlinks,bookmarks=false]{hyperref}

\begin{document}
\pagestyle{headings}
\mainmatter
\def\ECCVSubNumber{6959}  

\title{Behind the Scene: Revealing the Secrets of Pre-trained Vision-and-Language Models} 

\titlerunning{Revealing the Secrets of Pre-trained Vision-and-Language Models}
%
\author{Jize Cao\thanks{This work was done when Jize and Licheng worked at Microsoft.}\inst{1} \and
Zhe Gan\inst{2} \and Yu Cheng\inst{2} \and Licheng Yu$^\star$\inst{3} \\ Yen-Chun Chen\inst{2} \and 
Jingjing Liu\inst{2}}
\authorrunning{J. Cao et al.}
%
\institute{University of Washington \\
\email{caojize@cs.washington.edu} \and
Microsoft Dynamics 365 AI Research \\
\email{\{zhe.gan,yu.cheng,yen-chun.chen,jingjl\}@microsoft.com}
\and
Facebook AI\\
\email{lichengyu@fb.com}
}
\maketitle
\setcounter{secnumdepth}{3}

\begin{abstract}
Recent Transformer-based large-scale pre-trained models have revolutionized vision-and-language (V+L) research. Models such as ViLBERT, LXMERT and UNITER have significantly lifted state of the art across a wide range of V+L benchmarks. However, little is known about the inner mechanisms
that destine their impressive success. To reveal the secrets behind the scene, we present \textsc{Value} (\textbf{V}ision-\textbf{A}nd-\textbf{L}anguage \textbf{U}nderstanding \textbf{E}valuation), a set of meticulously designed probing tasks (\emph{e.g.}, Visual Coreference Resolution, Visual Relation Detection) generalizable to standard pre-trained V+L models, to decipher the inner workings of multimodal pre-training (\emph{e.g.}, implicit knowledge garnered in individual attention heads, inherent cross-modal alignment learned through contextualized multimodal embeddings). Through extensive analysis of each archetypal model architecture via these probing tasks, our key observations are: ($i$) Pre-trained models exhibit a propensity for attending over text rather than images during inference. ($ii$) There exists a subset of attention heads that are tailored for capturing cross-modal interactions. ($iii$) Learned attention matrix in pre-trained models demonstrates patterns coherent with the latent alignment between image regions and textual words. ($iv$) Plotted attention patterns reveal visually-interpretable relations among image regions. ($v$) Pure linguistic knowledge is also effectively encoded in the attention heads. These are valuable insights serving to guide future work towards designing better model architecture and objectives for multimodal pre-training.\footnote{Code is available at \url{https://github.com/JizeCao/VALUE}.}
\end{abstract}

\section{Introduction}
Recently, Transformer-based~\cite{transformer} large-scale pre-trained models~\cite{tan2019lxmert,lu2019vilbert,chen2019uniter,li2019unicoder,li2019visualbert,su2019vl} have prevailed in Vision-and-Language (V+L) research, an important area that sits at the nexus of computer vision and natural language processing (NLP). Inspired by BERT~\cite{bert}, a common practice for pre-training V+L models is to first encode image regions and sentence words into a common embedding space, then use multiple Transformer layers
to learn image-text contextualized joint embeddings through well-designed pre-training tasks. There are two main schools of model design: ($i$) single-stream architecture, such as VLBERT~\cite{su2019vl} and UNITER~\cite{su2019vl}, where a single Transformer is applied to both image and text modalities; and ($ii$) two-stream architecture, such as LXMERT~\cite{tan2019lxmert} and ViLBERT~\cite{lu2019vilbert}, in which two Transformers are applied to images and text independently, and a third Transformer is stacked on top for later fusion. When finetuned on downstream tasks, these pre-trained models have achieved new state of the art on image-text retrieval~\cite{lee2018stacked}, visual question answering~\cite{antol2015vqa,goyal2017making}, referring expression comprehension~\cite{yu2016modeling}, and visual reasoning~\cite{hudson2019gqa,suhr2018corpus,zellers2019recognition}. This suggests that substantial amount of visual and linguistic knowledge has been encoded in the pre-trained models. 

\begin{figure}[t!]
    \centering
    \includegraphics[width=\linewidth]{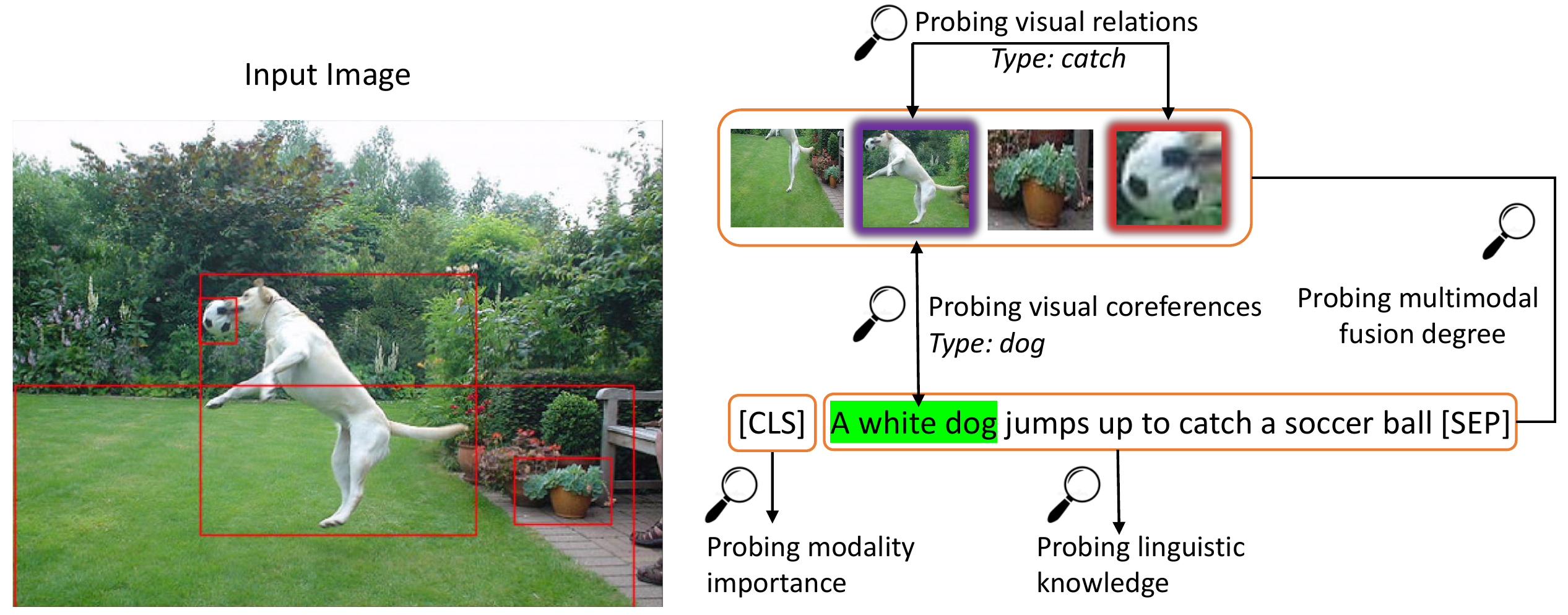}
    \caption{\label{fig:framework}\small{Illustration of the proposed \textsc{Value} framework for investigating pre-trained vision-and-language models. \textsc{Value} consists of a set of well-designed probing tasks that unveil the inner mechanisms of V+L pre-trained models across: ($i$) Multimodal Fusion Degree; ($ii$) Modality Importance; ($iii$) Cross-modal Interaction via probing visual coreferences; ($iv$) Image-to-image Interaction via probing visual relations; and ($v$) Text-to-text Interaction via probing learned linguistic knowledge.}}
\end{figure}

There has been several studies that investigate latent knowledge encoded in pre-trained language models~\cite{tenney2019bert,clark2019does,petroni2019language,kovaleva2019revealing}. However, analyzing multimodal pre-trained models is still an unexplored territory. It remains unclear how the inner mechanisms of cross-modal pre-trained models induce their empirical success on downstream tasks.
Motivated by this, we present \textsc{Value} (\textbf{V}ision-\textbf{A}nd-\textbf{L}anguage \textbf{U}nderstanding \textbf{E}valuation), a set of well-designed probing tasks that aims to reveal the secrets of these pre-trained V+L models. To investigate both single- and two-stream model architectures, we select one model from each category (LXMERT for two-stream
and UNITER for single-stream, because of their superb performance across many V+L tasks).
As illustrated in Figure~\ref{fig:framework}, \textsc{Value} is designed to provide insights on: ($a$) Multimodal Fusion Degree; ($b$) Modality Importance; ($c$) Cross-modal Interaction (Image-to-text/Text-to-image); ($d$) Image-to-image Interaction; and ($e$) Text-to-text Interaction. 

For ($a$) Multimodal Fusion Degree, clustering analysis between image and text representations shows that in single-stream models like UNITER, as the network layers go deeper, the fusion between two modalities becomes more intertwined. However, the opposite phenomenon is observed in two-stream models like LXMERT. For ($b$) Modality Importance, by analyzing the attention trace of the \texttt{[CLS]} token, which is commonly considered as containing the intended fused multimodal information and often used as the input signal for downstream tasks, we find that the final predictions tend to depend more on textual input rather than visual input.

To gain deeper insights into how pre-trained models drive success in downstream tasks, we look into three types of interactions between modalities. For ($c$) Cross-modal Interaction, we propose a Visual Coreference Resolution task to probe its encoded knowledge.
For ($d$) Image-to-image Interaction, we conduct analysis via Visual Relation Detection between two image regions. For ($e$) Text-to-text Interaction, we evaluate the linguistic knowledge encoded in each layer of the tested model with SentEval tookit~\cite{conneau2018senteval}, and compare with the original BERT~\cite{bert}. Experiments show that both single- and two-stream models, especially the former, can well capture cross-modal alignment, visual relations, and linguistic knowledge. 

To the best of our knowledge, this is the first known effort on thorough analysis of pre-trained V+L models, to gain insights from different perspectives about the latent knowledge encoded in self-attention weights, and to distill the secret ingredients that drive the empirical success of prevailing V+L models. 

\section{Related Work}

For image-text representation learning,
ViLBERT~\cite{lu2019vilbert} and LXMERT~\cite{tan2019lxmert} used two-stream architecture for pre-training, while B2T2~\cite{alberti2019fusion}, VisualBERT~\cite{li2019visualbert}, Unicoder-VL~\cite{li2019unicoder}, VL-BERT~\cite{su2019vl} and UNITER~\cite{chen2019uniter} adopted single-stream architecture. VLP~\cite{zhou2019unified} proposed a unified pre-trained model for both image captioning and VQA. Multi-task learning~\cite{lu201912} and adversarial training in VILLA~\cite{gan2020large} have been studied to boost performance. On video+language side, VideoBERT~\cite{sun2019videobert} applied BERT to learn joint embeddings of video frame tokens and linguistic tokens from video-text pairs. CBT~\cite{sun2019contrastive} introduced contrastive learning to handle real-valued video frame features, and HERO~\cite{li2020hero} proposed hierarchical Transformer architectures to leverage both global and local temporal visual-textual alignments. 
However, except for some simple visualization of the learned attention maps ~\cite{tan2019lxmert}, no existing work has systematically analyzed these pre-trained models. 

There has been some recent studies on assessing the capability of BERT in capturing structural properties of language~\cite{tenney2019you,jiang2019can,talmor2019olmpics}. Multi-head self-attention has been analyzed for machine translation in~\cite{voita2019analyzing,michel2019sixteen}, which observed that only a small subset of heads is important, and the other heads can be pruned without affecting model performance. 
\cite{tenney2019bert} reported that BERT can rediscover the classical NLP pipeline, where basic syntactic information appears in lower layers, while high-level semantic information appears in higher layers. Analysis on BERT self-attention~\cite{clark2019does,htut2019attention,kovaleva2019revealing} showed that BERT can learn syntactic relations, and a limited set of attention patterns are repeated across different heads. \cite{petroni2019language,bouraoui2019inducing} and \cite{zhou2019evaluating} demonstrated that BERT has surprisingly strong ability to recall factual relational knowledge and perform commonsense reasoning. A layer-wise analysis of Transformer representations in~\cite{van2019does} provided insights to the reasoning process of how BERT answers questions.
All these studies have focused on the analysis of BERT, while investigating pre-trained V+L models is still an uncharted territory. Given their empirical success and unique multimodal nature, we believe it is instrumental to conduct an in-depth analysis to understand these models, to provide useful insights and guidelines for future studies. New probing tasks such as visual coreference resolution and visual relation detection are proposed for this purpose, which can lend insights to other evaluation tasks as well.

\section{V\small{ALUE}: \large{Probing Pre-trained V+L Models}} \label{sec:probing_tasks}
Key curiosities this study aims to unveil include:

($i$) What is the correlation between multimodal fusion and the number of layers in pre-trained models? (Sec.~\ref{sec:modality_distinctiveness})
   
($ii$) Which modality plays a more important role that drives the pre-trained model to make final predictions? (Sec.~\ref{sec:modality_importance})

($iii$) What knowledge is encoded in pre-trained models that supports cross-modal interaction and alignment? (Sec.~\ref{sec:cross_modal_interactions})
    
($iv$) What knowledge has been learned for image-to-image (intra-modal) interaction (\emph{i.e.}, visual relations)? (Sec.~\ref{sec:visual_relations})

($v$) Compared with BERT, do pre-trained V+L models effectively encode linguistic knowledge for text-to-text (intra-modal) interaction? (Sec.~\ref{sec:lingustic_probing})

To answer these questions, we select one model from each archetypal model architecture for dissection: UNITER-base~\cite{chen2019uniter} (12 layers, 768 hidden units per layer, 12 attention heads) for single-stream model, and LXMERT~\cite{tan2019lxmert} for two-stream model.\footnote{Our probing analysis can be readily extended to other pre-trained models as well.}
Single-stream model (UNITER) shares the same structure as BERT \cite{bert}, except that the input now becomes a mixed sequence of two modalities. Two-stream model (LXMERT) first performs self-attention through several layers on each modality independently, then fuses the inputs through a stack of cross-self-attention layers (first cross-attention, then self-attention). Therefore, the attention pattern in two-stream models is fixed, as one modality is only allowed to attend over either itself or the other modality at any time (there is no such constraint in single-stream models). A more detailed model description is provided in the Appendix.

Two datasets are selected for our probing experiments:
 
($i$) \textbf{Visual Genome (VG)}~\cite{krishna2017visual}: image-text dataset with annotated dense captions and scene graphs. 
   
($ii$) \textbf{Flickr30k Entities}~\cite{flickr}: image-text dataset with annotated visual co-reference links between image regions and noun phrases in the captions.
   
Each data sample consists of three components: ($i$) an input image; ($ii$) a set of detected image regions;\footnote{An image region is also called a visual token in this paper; these two terms will be used interchangeable throughout the paper.} and ($iii$) a caption. In Flickr30k Entities, the caption is relatively long, describing the whole image; while in VG, a few short captions are provided (called dense captions), each describing an image region. 

The number of annotated image regions in an image is relatively small (5-6 for Flick30k Entities, and 2-4 for VG dense annotated graph); while in pre-trained V+L models, the number of image regions fed to the model is typically 36~\cite{BottomUpAT}. Therefore, we extract an additional set of image regions from a Faster R-CNN pre-trained on the VG dataset \cite{BottomUpAT}, and combine them with the original image regions provided in the dataset. The initial image representation is obtained from the Faster R-CNN as well. Finally, we feed both image regions and textual tokens into the pre-trained model for probing analysis. The following sub-sections describe analysis results and key observations from each proposed probing task.

\subsection{Deep to Profound: Deeper Layers Lead to More Intertwined Multimodal Fusion} \label{sec:modality_distinctiveness}
By nature, visual features in images and linguistic clues in text have distinctive characteristics. However, it is unknown whether the semantic gap between their corresponding representations narrows (\emph{i.e.}, the contextualized representations of the two modalities become less differentiable) through the cross-modal fusion between intermediate layers.  
\subsubsection{Probing Task}
To answer this question, we design a probing task to test the multimodal fusion degree of a model. First, we extract all the embedding features of both image regions and textual tokens from aforementioned two datasets (VG and Flickr30k Entities). For single-stream model (UNITER-base), we extract the output representation from each layer, as multimodal fusion is performed through all the layers. For two-stream model (LXMERT), we only consider the output embeddings from the last 5 layers (\emph{i.e.}, the layers constituting the cross-modality encoder), because the two modalities do not have any interaction prior to that. 
To gain quantitative measurement, for each data sample, we apply the $k$-means algorithm (with $k$ = 2) on the representations from each layer to partition them into two clusters, and measure the difference between the formed clusters and ground-truth visual/textual clusters via Normalized Mutual Information (NMI),
an unsupervised metric for evaluating differences between clusters. 
A larger NMI value implies that the distinction between two clusters is more significant, indicating a lower fusion degree~\cite{jawahar2019does}.
For example, when NMI is equal to 1.0, the two clusters represent the original visual tokens and textual tokens, respectively.
We use the mean value of NMI for all the data samples to measure the level of multimodal fusion. 

\input{tables/nmi_scores}
\subsubsection{Results}
Table~\ref{tab:modality_distinctiveness} summarizes the probing results on multimodal fusion degree. For single-stream model (UNITER-base), the NMI scores gradually decrease, indicating that the representations from the two modalities fuse together deeper and deeper from lower to higher layers. 
This observation matches our intuition that the embedding of a modality from a higher layer better attends over the other modality than a lower layer. However, in two-stream model (LXMERT), as the layers go deeper (except for the last layer), the representations deviate from each other. 
One possible explanation is that single-stream model applies the same set of parameters to both image and text modalities, while two-stream model uses two separate sets of parameters (as part of its network design) to model the attention on the image and text modality independently. 
The latter makes it relatively easier to distinguish the two representations, leading to a higher NMI score. An t-SNE visualization is provided in the Appendix. 

\subsection{Who Pulls More Strings: Textual Modality Is More Dominant than Image} \label{sec:modality_importance}

Following BERT~\cite{bert}, pre-trained V+L models add a special \texttt{[CLS]} token at the beginning of a sequence, and a special \texttt{[SEP]} token at the end. In practice,  the \texttt{[CLS]} token representation from the last layer is used as the fused representation for both modalities in downstream tasks. Since the \texttt{[CLS]} token absorbs information from both modalities through self-attention, the degree of attention of the \texttt{[CLS]} token over each modality could be regarded as evidence to answer the following question: \emph{which modality is more dominant during inference?} Note that in the two-stream model design, the \texttt{[CLS]} token is not allowed to attend to both image and textual modality simultaneously. Therefore, the probing experiments for modality importance is focused on single-stream model. 

\begin{figure}[t!]
\centering
\begin{subfigure}[b]{5.25cm}            
\includegraphics[width=5cm]{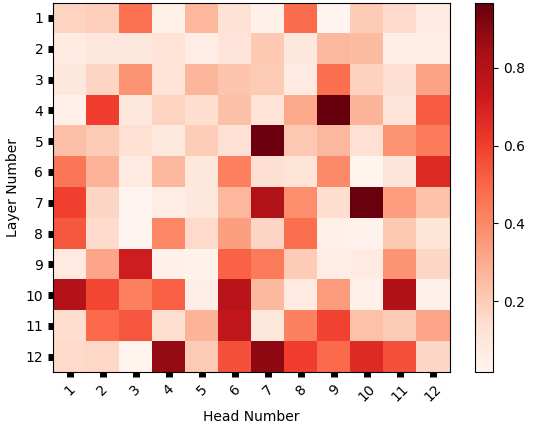}
\caption{\small{Textual modality importance}}
\end{subfigure}
\begin{subfigure}[b]{5.25cm}
\centering
\includegraphics[width=5cm]{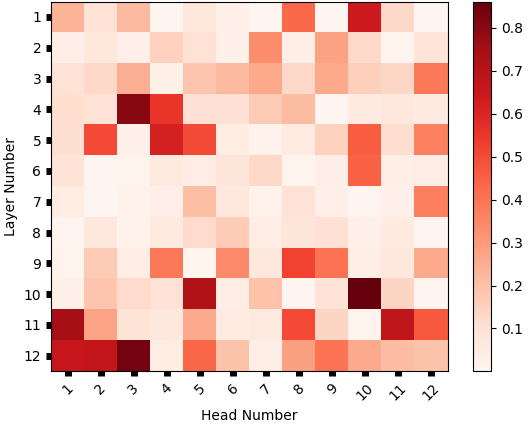}
\caption{\small{Visual modality importance}}
\end{subfigure}
\caption{\label{fig:head_wise_modality_importance}\small{Visualization of the modality importance score for all the 144 attention heads.}}
\end{figure}

\subsubsection{Probing Task}
Formally, to quantitatively analyze the \texttt{[CLS]} attention trace, the \textit{Modality Importance} (MI) of a head $j$ is defined as the sum of the attention values that the \texttt{[CLS]} token spent on the modality $M$ (visual or textual) at $j$  for the whole sequence $S =$ (\texttt{[CLS]}, $t_1,\ldots,t_m$, \texttt{[SEP]}, $v_1,\ldots,v_n$), where $t_1,\ldots,t_m$ and $v_1,\ldots,v_n$ are textual and visual tokens, respectively. That is, 
\begin{align}
  I_{M,j} = \textstyle{\sum_{i \in S}} \mathbbm{1}(i \in M) \cdot \alpha_{ij}\,,
\end{align}
where $\alpha_{ij}$ represents the attention weight that the \texttt{[CLS]} token attends on token $i$ at head $j$, and $\mathbbm{1}(\cdot)$ is the indicator function.

Since the attention heads in the same layer perform attention on the same representation, it is natural to expand the above head-level analysis to a layer-level analysis, by considering the mean of MI scores of all the 12 heads as the MI measurement of that layer.
In addition, we calculate an overall MI score via summing up the MI scores of all the 144 heads. The mean MI value of all the data samples is reported as the score for probing modality importance. Note that the MI score for visual/textual modality is calculated based on the attention weights on the visual/textual tokens; while the attention weights on the special \texttt{[CLS]} and \texttt{[SEP]} tokens are not considered. Therefore, the two mean MI scores from the two modalities do not sum to one. 

\subsubsection{Results}
\begin{figure}[t!]
\centering
\begin{subfigure}[b]{5.5cm}            
\includegraphics[height=4.5cm]{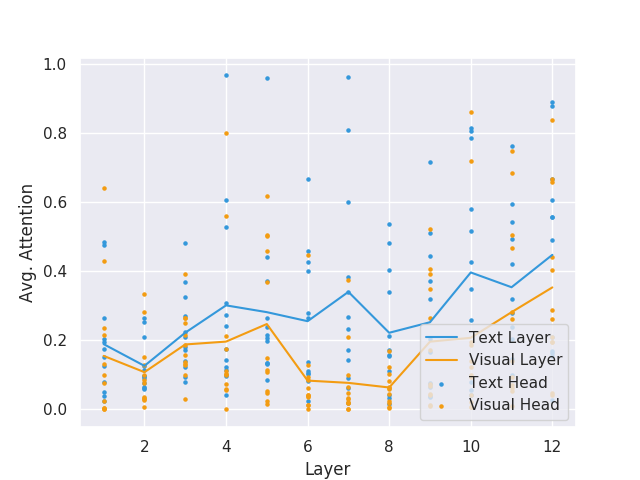}
\caption{\label{fig:modality_importance}\small{Layer-level modality importance}}
\end{subfigure}
\begin{subfigure}[b]{5.5cm}
\centering
\includegraphics[width=5cm]{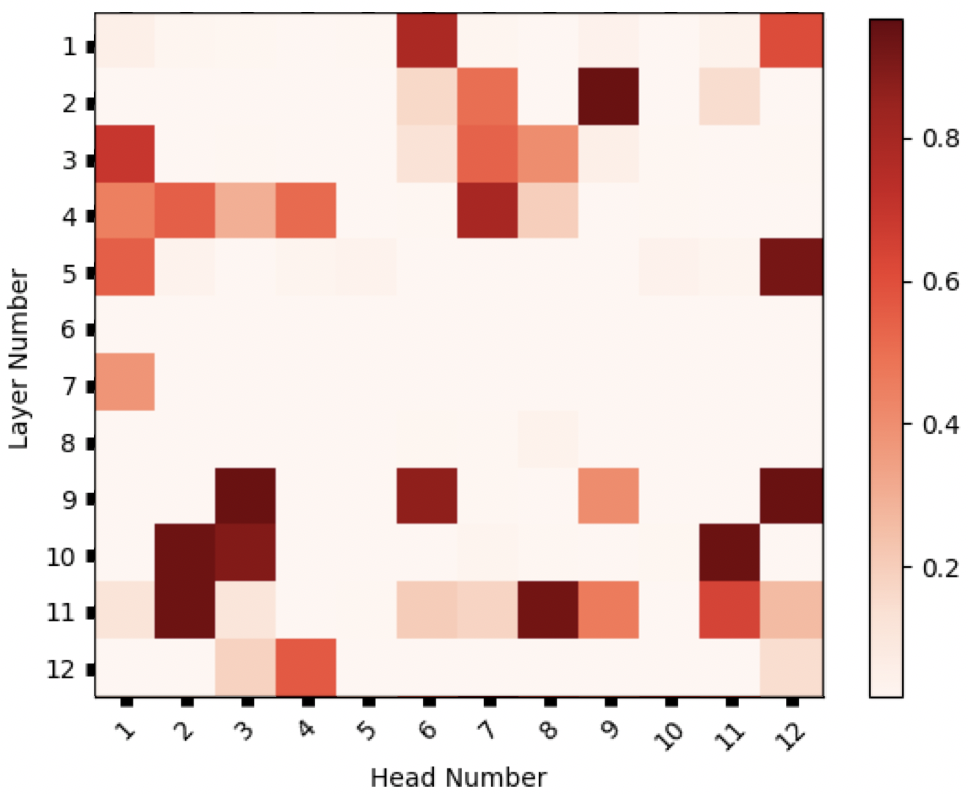}
\caption{\label{fig:image_text_focus_head} \small{Image-to-text attention}}
\end{subfigure}
\caption{\small{Layer-level modality importance scores and visualization of the image-to-text attention for all the 144 attention heads.}}
\end{figure}

Experiments are conducted on the Flickr30k Entity dataset.
Figure~\ref{fig:head_wise_modality_importance} provides the MI scores for all the 144 attention heads of UNITER-base. 
The heatmap on the textual MI is denser than that of the visual MI, showing that more attention heads are learning useful knowledge from the textual modality than the image modality. 
Figure~\ref{fig:modality_importance} further shows the layer-level MI scores for each modality. The average MI score on the text modality is higher than that on the image modality, especially for intermediate layers, suggesting that the pre-trained model relies more on the textual modality for making decisions during inference time.  

\subsection{Winner Takes All: A Subset of Heads is Specialized for Cross-modal Interaction} \label{sec:cross_modal_interactions}

The key difference between single-modal Transformer (such as BERT) and two-modal Transformer (such as UNITER and LXMERT) is that two-modal Transformer requires extra cross-modal interaction. To gain an in-depth understanding of cross-modal attention heads, which is instructive to prompt better model design and enhance model interpretability, we look into two special types of head: ($i$) image-to-text head; and ($ii$) visual-coreference head.

\subsubsection{Probing Image-to-text Head} \,
We first analyze whether there exists any head specialized in learning cross-modal interaction. Formally, for a given image-text pair, visual and textual tokens are denoted as $V$ and $T$, respectively. We define a head as \textit{image-to-text head} if: 
\begin{align}
    \exists \, v \in V \, \textstyle{\sum_{t \in T}} \, \alpha_{v \to t} > 0.5 \,,
\end{align}
where $\alpha_{v\to t}$ denotes the attention weight from a visual token to a textual token.
This defines whether a visual token pays more attention to the text modality. Specifically, if there exists one visual token that has higher attention weight on text than other tokens, we regard the corresponding head as performing cross-modal attention from image to text. 

Based on the above definition, we count the number of occurrences of head as an \emph{image-to-text head} for all data samples in the Flickr30k Entities dataset, and report the empirical probability of each head being an \emph{image-to-text head}. Note that in the two-stream model, the image-to-text head is by design, therefore, we only conduct this analysis on single-stream model.

Figure~\ref{fig:image_text_focus_head} shows there is a specific set of heads in UNITER-base that perform cross-modal interaction. 
The maximum probability of a head performing image-to-text attention is 0.92, the minimum probability is 0, and 
only 15\% heads have more than 0.5 probability to pay the majority attention weight on the image-to-text part.
Interestingly, by training single-stream model, the attention heads are automatically learned to exhibit a ``two-stream" pattern, where some heads control the message sharing from the visual modality to the textual modality.

\subsubsection{Visual Coreference Resolution} \,
One straightforward way to investigate the visual-linguistic knowledge encoded in the model is to evaluate whether the model is able to match an image region to its corresponding textual phrase in the sentence. Thus, we design a \emph{Visual Coreference Resolution} task (similar to coreference resolution) to predict whether there is a link between an image region and a noun phrase in the sentence that describes the image. In addition, each coreference link in the dataset is annotated with a label (8 in total). 

Through this task, we can find out whether the coreference knowledge can be captured by the attention trace. To achieve this goal, for each data sample in the Flickr30k Entity dataset, we extract the encoder's attention weights for all the 144 heads. Note that noun phrases typically consist of two or more tokens in the sequence. Thus, we extract the maximum attention weight between the image region and each word of the noun phrase for each head. The maximum weight is then used to evaluate which head identifies visual coreference (\emph{i.e.}, performing multimodal alignment).

\input{tables/vcr_results_1}

Results are summarized in Table~\ref{tab:coref_resolution}. The columns labeled with ``(Rand.)'' are considered as ablation groups to identify whether the high attention weight of a certain coreference relationship is triggered by the relation between the image-text pair rather than the effect of one specific image/text token. For a link $V \to T$, we measure the maximum attention weight of the visual token $V$ to a random noun phrase, to obtain the results for $V \to T$ (Rand.). Similarly, we use the maximum attention weight of a noun phrase $T$ to a random visual token to obtain the results for $T \to V$  (Rand.).

Results show that the relation between a noun phase and its linked visual token is encoded in the attention pattern, especially for $V \to T$. The heads (9-3)\footnote{Head ($i$-$j$) means the $j$-th head at the $i$-th layer.}, (9-12) and (3-1) have captured richer coreference knowledge than other heads, indicating that there exists a subset of heads in the pre-trained model that is specialized in coreference linking between the two modalities. Furthermore, for single-stream model, we observe that some heads encode the coreference information in both directions, serving as additional evidence that these heads perform cross-modal alignment specifically.
On the other hand, the amount of learned coreference knowledge is limited in the text modality attention trace ($T\to V$), which provides indirect evidence that the text modality does not incorporate much visual information, even with the forced cross-attention design in the two-stream model.


\subsubsection{Probing Combinations of Heads} \,
Previous analysis mainly investigates whether cross-modal knowledge can be captured through individual attention head. It is also possible that such knowledge can be induced via the cooperation of multiple heads. To quantitatively analyze this, we further examine visual coreference through probing over combinations of heads. 

\textbf{1) Attention Prober} \, 
To reveal the learned knowledge across different attention heads, we use a linear classifier based on the combination of attention weights, following~\cite{clark2019does}. Specifically,
\begin{align} \label{eqn:att_prober}
    p(c | i, j) \propto \textstyle{\sum_{k = 1}^N} (w_k \alpha_{ij}^k + \mu_k\alpha_{ji}^k) \,,
\end{align}
where given tokens $i$ and $j$ in the sequence, $p(c | i, j)$ is the probability of the link label between these two tokens being $c$. $\alpha_{ij}^k$ is the attention weight for token $i$ attending to token $j$ at head $k$, $w_k$ and $\mu_k$ are two learnable scalars, and $N$ is the number of attention heads.

\textbf{2) Layer-wise Embedding Prober} \, 
Similarly, to further examine the knowledge encoded in the model, we can naturally extend the above attention-based prober into an embedding-based prober: 
\begin{align}
    p_k(c | i, j) \propto w_{cls}^{k\top} (W_{w}^k e_{i}^k \odot W_{\mu}^k e^k_{j}) \,,
\end{align}
where $p_k(c| i, j)$ is the same as defined in (\ref{eqn:att_prober}), $e_i^k$ and $e_j^k$ are the embeddings of token $i$ and $j$ at the $k$-th layer, and $W_{w}^k$, $W_{\mu}^k \in {\mathbb R}^{768 \times 768}, w_{cls}^k \in {\mathbb R}^{768 \times 1}$. By training a linear classifier on top of extracted embeddings, this prober provides a way to probe the encoded knowledge between each pair of tokens on each layer.

For visual coreference resolution, we probe the model on two sub-tasks: ($i$) \texttt{Visual Coref Detection (VCD)}: determine whether a noun phrase is coreferenced to a specific visual token (binary classification); and ($ii$) \texttt{Visual Coref Classification (VCC)}: classify the label of the coreference relation between a noun phrase and a visual token (multi-class classification). With these tasks, we can examine whether certain type of knowledge is encoded, as well as the granularity of the knowledge encoded in the probed feature space\footnote{Since noun phrase may contain several tokens, we use the maximum attention weight among the tokens in that phrase over an image region as the attention weight between the noun phase and the image region. The embedding of the noun phrase is the mean of all the representations of its textual tokens.}.

\input{tables/vcr_results_2}

Results are shown in Table~\ref{tab:visual_coreference}. 
Neither model performs well on the VCD task, suggesting that there is no attention pattern or embedding feature that can handle all the coreference relations. 
However, the results on VCC are encouraging. This aligns with our observation from Table~\ref{tab:coref_resolution} that some attention heads of the two models are significantly effective in certain coreference relations.\footnote{Though both models' embedding probers achieve higher than 94\% accuracy on the VCC task, it is worth noting that text embedding input can potentially leak the link information.
For instance, the phrase ``A guard with a white hat" may already provide coreference information between \texttt{person} and the corresponding image region.}

\subsection{Secret Liaison Revealed: Cross-Modality Fusion Registers Visual Relations} \label{sec:visual_relations}

To evaluate the encoded knowledge learned from the image modality, we adopt the visual relation detection task, which requires a model to identify 
and classify the relation between two image regions. This task can be viewed as examining whether the model captures visual relations between image regions. 

\input{tables/vrd_results}

The VG dataset is used for this task, which contains 1,531,448 first-order object-object relations. First-order object-object relation can be determined simply by the visual representations of two objects, independent to other objects in the image or text annotation. To reduce the imbalance in the number of relations per relation type, we randomly select at most 15,000 subject-object relation pairs $(s, o)$ per relation type. Furthermore, to de-duplicate cases where the same type of relation comes from the same text annotation, we select at most 5 same relation types $(s, o)$ from the same annotation. This probing task is performed on 32 most frequent relation pairs in the dataset. 

\textbf{1) Probing Individual Attention Head} \, 
We apply the same analysis here similar to the visual coreference resolution task. The only difference is that we do not consider the directions of attention (\emph{i.e.}, $s \to o$, or $o \to s$), as both directions correspond to the same visual modality. The average of maximum attention value in both $s \to o$ and $o \to s$ directions is reported for visual relation analysis.

Results are summarized in Table~\ref{tab:visual_relation}. We report the maximum attention weights of 7 of 30 relations. The columns with ``(Rand.)'' are ablation groups. As shown in the comparison, the learned attention heads encode rich knowledge about the relations between visual tokens with much higher attention weights. Moreover, similar to the observation in visual coreference resolution task, specific heads (10-1 head in the single-stream model, 7-4 head in the two-stream model) have captured richer visual relations than others.

\textbf{2) Probing Combinations of Attention Heads} \, 
The above analysis only reveals the behavior of individual attention heads. Similar to Sec.~\ref{sec:cross_modal_interactions}, we further examine visual relations by training a linear classifier on top of a combination of attention heads over two sub-tasks: ($i$) \texttt{Relation Identification}: determine whether two image regions have a relationship; and ($ii$) \texttt{Relation Classification}: classify the relation label between two image regions.

\begin{figure}[t!]
\begin{subfigure}[b]{\textwidth}
\centering
\begin{tabular}{ccccc}
\toprule
\bf Classifier Input & \bf VRI (SS) & \bf VRC (SS) & \bf VRI (TS) & \bf VRC (TS) \\
\midrule
144 Attention Heads & \textbf{69.81} & \textbf{24.67}  & \textbf{67.53} & \textbf{18.89} \\
144 Attention Heads (mismatch) & 64.66 & 23.64 & 64.72 & 18.42 \\
Random Guess & 50.00 & 3.33  & 50.00 & 3.33 \\
\bottomrule
\end{tabular}
\caption{Accuracies (\%) of the attention probers on VRI and VRC}
\end{subfigure}
\begin{subfigure}[b]{5.85cm}
\centering
\includegraphics[width=5.5cm]{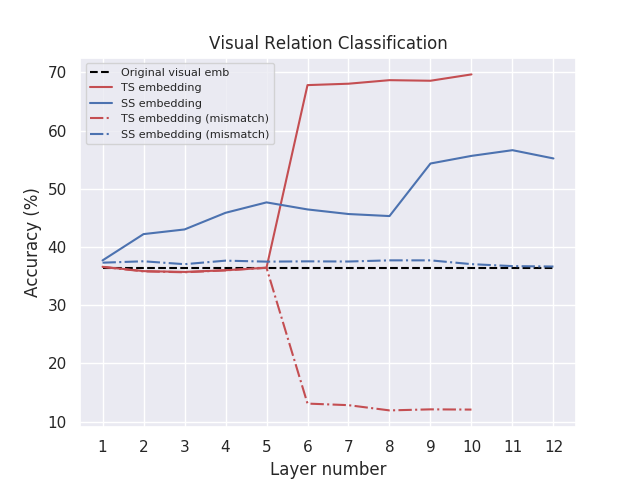}
\caption{\small{Layer-wise emb. prober on VRC}}
\end{subfigure}
\begin{subfigure}[b]{5.85cm}
\centering
\includegraphics[width=5.5cm]{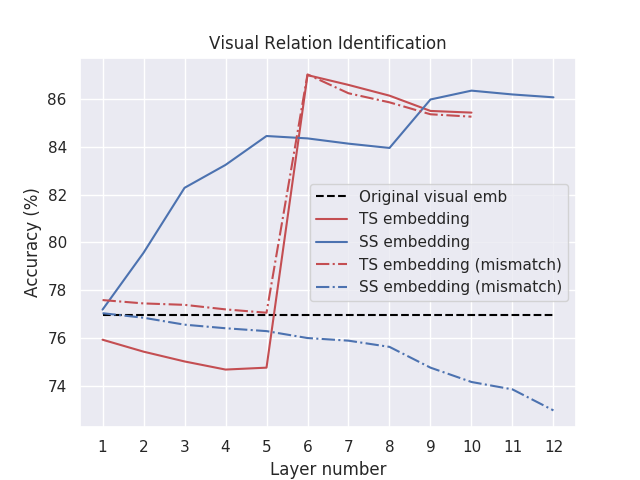}
\caption{\small{Layer-wise emb. prober on VRI}}
\end{subfigure}
\caption{\label{fig:visual_relation_probing}\small{Results of attention and layer-wise embedding probers on Visual Relation Identification and Classification (VRI and VRC)}.}
\end{figure}

Results on probing a combination of heads are summarized in Figure~\ref{fig:visual_relation_probing}.
Two baselines are considered in this task. ($i$) Original visual embeddings from Faster R-CNN \cite{BottomUpAT}. 
This setup evaluates how much correlation between two related visual representations has elevated or diluted by the V+L models. ($ii$) Mismatched image-text representation. In this baseline, we construct a dataset where an image is associated with an unrelated dense annotation instead of a related one. This baseline evaluates how the correlation between the image regions changes based on the text modality. Note that for the two-stream model, there are 10 layers involved in visual relation reasoning. The first five layers perform self-attention across the visual modality only, and in the last five layers the visual representation interacts with the text modality.

As shown, both models perform much better than the baseline with related annotations (the dashed balck line in Figure~\ref{fig:visual_relation_probing}), indicating there is a substantial amount of visual-relation knowledge encoded in these representations. On the other hand, the specific visual relation knowledge degrades a lot with mismatched image regions and dense annotations (the solid red and blue lines vs. the dashed lines in Figure~\ref{fig:visual_relation_probing}). 
For single-stream model, the visual-relation knowledge decayed to that of the original visual embedding. This makes sense because no visual relation information can be obtained from the mismatched caption. 

The visual relation knowledge of the two-stream model is greatly influenced by the unpaired caption, leading to a huge performance drop after Layer 5 in the VRC task. This result may contribute to different inductive bias between the two models. For two-stream model, since the visual modality has to attend to the text modality during cross attention, the visual representation will be greatly influenced by the text modality even when the two modalities are totally unrelated. On the other hand, because the visual representation in the single-stream model is capable of selectively choosing whether to attend over the text modality, the effect of unrelated caption on the visual representation is negligible.

\subsection{No Lost in Translation: Pre-trained V+L Models Encode Rich Linguistic Knowledge} \label{sec:lingustic_probing}

Besides looking into the knowledge learned from the visual modality, we are also interested in the encoded knowledge learned from the text modality. To achieve this goal, we probe the pre-trained models over nine tasks defined in the SentEval toolkit~\cite{conneau2018senteval}. Descriptions about the tasks are provided in the Appendix.

First, we extract contextualized word representations from the pre-trained models. For single-stream model, the input is the sequence of tokens with text only. For two-stream model, since the last 5 cross-attention layers require visual input, which is not covered by these linguistic benchmarks, we only evaluate the first 9 layers that are performing self-attention over pure text input.

\input{tables/linguistic_probe}

We use the Google-pretrained BERT-base model as the baseline. For each task, we obtain the results from all the layers and report the best number. The results in Table~\ref{tab:lingustic_probing} show that pre-trained V+L model generally performs worse than the original BERT-base model on these linguistic benchmarks, which is as expected. A full table with results from all the layers is provided in the Appendix. As shown in the table, the single-stream model performs better than the two-stream model across all the tasks. 
One possible reason is that LXMERT does not initialize the parameters of the language Transformer encoder from BERT-base, whereas UNITER does. Thus, using the parameters of BERT as initialization is potentially useful for the model to acquire rich linguistic knowledge, and helpful for tasks involving complex text-based reasoning.

To measure the gains due to learning, we have conducted all the above experiments (Secs.~\ref{sec:modality_distinctiveness}-\ref{sec:lingustic_probing}) on untrained baselines with random weights. These additional results are provided in the Appendix.

\section{Conclusion and Key Takeaways} \label{sec:conclusion}

Intrigued by the five questions presented at the beginning of Section~\ref{sec:probing_tasks}, we have provided a thorough analysis of UNITER-base and LXMERT models as a deep dive into Vision+Language pre-training. 
To summarize our key findings: 

\vspace{5pt}
($i$) In single-stream model, deeper layers lead to more intertwined multimodal fusion; while the opposite trend is observed in two-stream model.

($ii$) Textual modality plays a more important role than image in making final decisions, consistent across both single- and two-stream models.

($iii$) In single-stream model, a subset of heads organically evolves to pivot on cross-modal interaction and alignment, which on the other hand is enforced by model design in two-stream model. 

($iv$) Visual relations are inherently registered in both single- and two-stream pre-trained models.

($v$) Rich linguistic knowledge is naturally encoded, even though the models are specifically designed for multimodal pre-training.

We provide additional guidelines in the Appendix.
For future work, we plan to perform model compression via pruning attention heads based on the analysis and observations in this work. 

\clearpage
\bibliographystyle{splncs04}
\bibliography{egbib}

\clearpage
\input{supp}

\end{document}

%% file: tables/nmi_scores.tex
\setlength{\tabcolsep}{2pt}
\begin{table}[t!]
\begin{center}
\begin{tabu}{lllllllllllll}
\Xhline{2\arrayrulewidth}
\bf Layer & \bf0 &  \bf1&  \bf2 &\bf 3 &\bf 4 &\bf 5 &\bf 6 &\bf 7 &\bf 8 & \bf 9 &\bf 10 &\bf 11 \\
\hline
\rowfont{\small}
\emph{single-stream} \\
Flickr30k & 0.36 & 0.38 & 0.39 & 0.41 & 0.38 & 0.38 & 0.38 & 0.38 & 0.32 & \textbf{0.20} & 0.26 & \textbf{0.20} \\
Visual Genome & 0.25 & 0.25 & 0.24 & 0.24 & 0.22 & 0.22 & 0.21 & 0.21 & 0.20 & 0.17 & \textbf{0.16} & \textbf{0.16} \\
\hline 
\rowfont{\small}
\emph{two-stream (cross)} \\
Flickr30k & \textbf{0.42} & 0.48 & 0.67 & 0.75 & 0.43 & $-$ & $-$ & $-$ & $-$ & $-$ & $-$ & $-$\\
Visual Genome & \textbf{0.43} & 0.56 & 0.68 & 0.78 & 0.57 & $-$ & $-$ & $-$ & $-$ & $-$ & $-$ & $-$ \\
\Xhline{2\arrayrulewidth}
\end{tabu}
\end{center}
\caption{\label{tab:modality_distinctiveness}\small{NMI scores on multimodal fusion probing. A smaller NMI value indicates a higher fusion degree. Note that the two-stream model (LXMERT) only has 5 layers in its cross-modality encoder. A larger layer number corresponds to an upper layer.}}
\end{table}
\setlength{\tabcolsep}{1.4pt}

%% file: tables/vcr_results_1.tex
\setlength{\tabcolsep}{2pt}
\begin{table}[t!]
\begin{center}
\begin{tabu}{lllll}
\Xhline{2\arrayrulewidth}
\multicolumn{3}{l} {\emph{Results on single-stream model}} \\
\hline
\bf Coref Type & $V \to T$   & $T \to V$ & $V \to  T$ (Rand.) & $T \to V$ (Rand.) \\
\hline
\tt people & \bf 0.165 (9-12) & 0.060 (3-1) & 0.106 (9-12) & 0.035 (11-8)  \\
\hline
\tt bodyparts & \bf 0.108 (9-3) & 0.051 (11-8) & 0.084 (9-12) & 0.038 (11-8) \\
\hline
\tt scene & \bf 0.151 (9-12) & 0.048 (3-1) & 0.111 (9-12) & 0.035 (11-8) \\
\hline
\tt clothing & \bf 0.157 (9-3) & 0.040 (3-1) & 0.092 (9-12) & 0.040 (10-2) \\
\hline
\tt animals & \bf 0.285 (9-12) & 0.137 (3-1) & 0.139 (9-12) & 0.047 (9-12) \\
\hline
\tt instruments & \bf 0.244 (11-8) & 0.042 (9-12) & 0.091 (9-12) & 0.031 (9-12)  \\
\hline
\tt vehicles & \bf 0.194 (9-12) & 0.065 (3-1) & 0.112 (9-12) & 0.039 (9-12) \\
\Xhline{2\arrayrulewidth}
\end{tabu} \\ 
\begin{tabu}{lllll}
\Xhline{2\arrayrulewidth}
\multicolumn{3}{l} {\emph{Results on two-stream model}} \\
\hline
\bf Coref Type & $V \to T$   & $T \to V$ & $V \to T$ (Rand.) & $T \to V$ (Rand.) \\
\hline
\tt people & \bf 0.145 (2-8) & 0.063 (2-6) & 0.056 (1-7) & 0.057 (3-9)  \\
\hline
\tt bodyparts & \bf 0.079 (5-7) & 0.059 (1-7) & 0.041 (1-2) & 0.060 (3-9) \\
\hline
\tt scene & \bf 0.076 (5-7) & 0.059 (1-7) & 0.038 (1-7) & 0.060 (3-9) \\
\hline
\tt clothing & \bf 0.062 (5-7) & 0.062 (2-6) & 0.040 (1-7) & 0.061 (3-9) \\
\hline
\tt animals & \bf 0.235 (3-4) & 0.106 (4-7) & 0.075 (1-7) & 0.072 (3-9) \\
\hline
\tt instruments & \bf 0.144 (5-7) & 0.040 (1-7) & 0.055 (1-11) & 0.058 (3-9)  \\
\hline
\tt vehicles & \bf 0.097 (1-1) & 0.046 (2-6) & 0.097 (1-7) & 0.062 (3-9) \\
\Xhline{2\arrayrulewidth}
\end{tabu}
\end{center}
\caption{\label{tab:coref_resolution}\small{Results on visual coreference resolution. Each number represents the maximum attention weights between two linked tokens, averaged across all data samples. $V \to T$ records the attention trace where a visual token attends to the linked noun phrase; $T \to V$ records the attention trace on the other direction. The head that achieves the maximum attention weight is shown in the bracket.}}
\end{table}
\setlength{\tabcolsep}{1.4pt}

%% file: tables/vcr_results_2.tex
\begin{table}[t!]
\centering
\begin{tabular}{ccccc}
\toprule
\bf Classifier Input & \bf VCD (SS) & \bf VCD (TS) & \bf VCC (SS) & \bf VCC (TS)  \\
\midrule
144 Attention Heads & 52.04  & \textbf{53.68} & 75.10 & 54.47 \\
Random Guess & 50.00 & 50.00 & 12.50 & 12.50 \\
\midrule
Layer 1 & 56.86 & \textbf{53.68} & 93.51 & \textbf{93.35}  \\
Layer 5 & \textbf{59.12} & 52.59 & \textbf{94.05} & 92.62 \\
Layer 12 & 58.40 & / & 93.44 & / \\
\bottomrule
\end{tabular}
\caption{\label{tab:visual_coreference}\small{Results of attention and layer-wise embedding probers on Visual Coreference Detection and Classification (VCD and VCC). SS: single-stream; TS: two-stream.}}
\end{table}

%% file: tables/vrd_results.tex
\setlength{\tabcolsep}{2pt}
\begin{table}[t!]
\begin{center}
\begin{tabu}{lcccc}
\Xhline{2\arrayrulewidth}
\bf Relation Type & \bf SS & \bf SS (Rand.) & \bf TS & \bf TS (Rand.) \\
\hline
\tt on & \bf 0.154 (10-1) & 0.055 (1-8) & \bf 0.157 (3-12) & 0.063 (5-9) \\
\hline
\tt standing in & \bf 0.107 (2-8) & 0.051 (1-8) & \bf 0.173 (7-4) & 0.064 (3-1) \\
\hline
\tt wearing & \bf 0.311 (10-1) & 0.049 (1-8) & \bf 0.230 (7-4) & 0.055 (3-1) \\
\hline
\tt playing & \bf 0.135 (4-1) & 0.050 (1-8) & \bf 0.103 (7-10) & 0.062 (3-1)  \\
\hline
\tt eating & \bf 0.138 (10-1) & 0.056 (1-8) & \bf 0.142 (7-4) & 0.067 (3-1) \\
\hline
\tt holds & \bf 0.200 (10-1) & 0.055 (1-8) & \bf 0.173 (7-4) & 0.066 (3-1) \\
\hline
\tt covering & \bf 0.151 (7-2) & 0.053 (1-8) & \bf 0.173 (3-1) & 0.061 (3-6) \\
\Xhline{2\arrayrulewidth}
\end{tabu} 
\end{center}
\caption{\label{tab:visual_relation}\small{Results on Visual Relation Identification/Classification using maximum attention weight between two visual tokens. SS: single stream; TS: two stream.}}
\end{table}
\setlength{\tabcolsep}{1.4pt}

%% file: tables/linguistic_probe.tex
\begin{table}[t!]
\centering
\resizebox{0.99\textwidth}{!}{%
\begin{tabular}{ c*{9}{|c}}
\toprule
\multicolumn{1}{l} {\bf Model}  & \multicolumn{1}{l} {\bf SentLen} & \multicolumn{1}{l} {\bf TreeDepth} & \multicolumn{1}{l} {\bf TopConst} & \multicolumn{1}{l} {\bf BShift} & \multicolumn{1}{l} {\bf Tense} & \multicolumn{1}{l} {\bf SubjNum} & \multicolumn{1}{l} {\bf ObjNum} & \multicolumn{1}{l} {\bf SOMO} & \multicolumn{1}{l} {\bf CoordInv} \\
 \multicolumn{1}{l} {} &   \multicolumn{1}{l} {(Surface)} &  \multicolumn{1}{l} {(Syntactic)} &  \multicolumn{1}{l} {(Syntactic)} &  \multicolumn{1}{l} {(Syntactic)} &  \multicolumn{1}{l} {(Semantic)} &  \multicolumn{1}{l} {(Semantic)} &  \multicolumn{1}{l} {(Semantic)} &  \multicolumn{1}{l} {(Semantic)} &  \multicolumn{1}{l} {(Semantic)} \\
\midrule

SS & 88.8 (L2) & 36.4 (L5) & 79.0 (L7) & 81.6 (L10) & 86.6 (L11) & 83.4 (L7) & 78.8 (L9) & 57.1 (L12) & 62.1 (L11) \\
TS & 83.8 (L7) & 34.0 (L8) & 67.2 (L8) &  64.9 (L8) & 75.6 (L6) & 78.8 (L9) & 76.8 (L8) & 51.4 (L8) & 58.7 (L8) \\
BERT & \bf 96.2 (L3) & \bf 41.3 (L6) & \bf 84.1 (L7) & \bf 87.0 (L9) & \bf 90.0 (L9) & \bf 88.1 (L6) & \bf 82.2 (L7) & \bf 65.2 (L12) & \bf 78.7 (L9) \\
\Xhline{2\arrayrulewidth} 
\end{tabular}}
\caption{\label{tab:lingustic_probing}\small{Results on the linguistic probing tasks. L$n$: Layer $n$.}}
\end{table}

%% file: supp.tex
\appendix 

\section{Details on Pre-trained V+L Models} \label{sec:pretrained_model_details}

\begin{figure}[h!]
    \centering
    \includegraphics[width=0.99\linewidth]{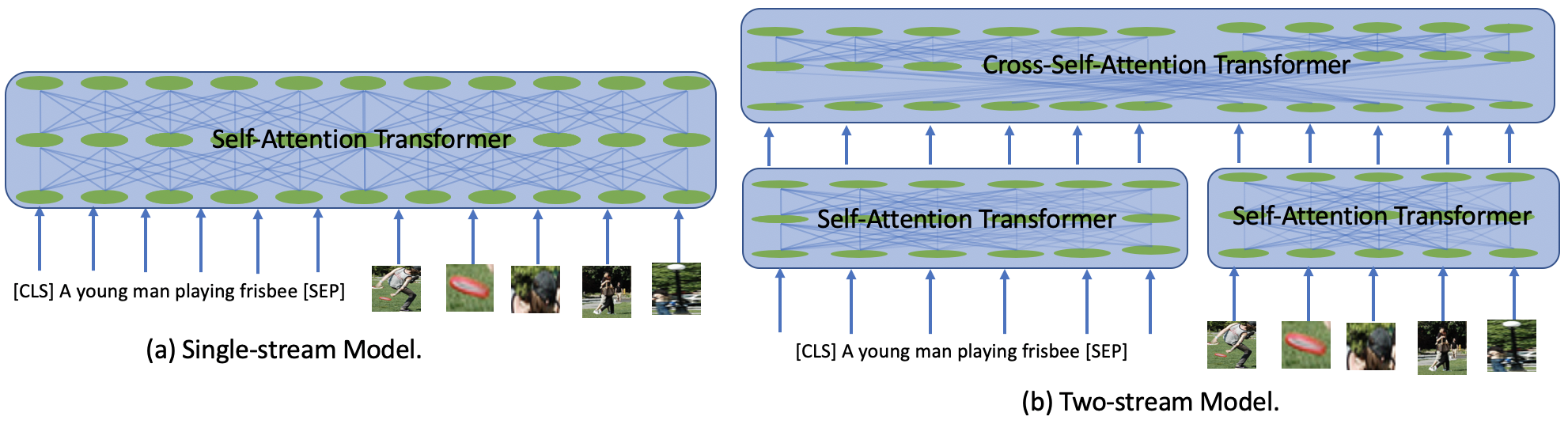}
    \caption{\label{fig:pretrained_model}\small{Comparison between single-stream and two-stream V+L models.}}
    \vspace{-2mm}
\end{figure}

A comparison between single-stream and two-stream V+L models is provided in Figure~\ref{fig:pretrained_model}. We choose UNITER-base\footnote{\url{https://github.com/ChenRocks/UNITER}} \cite{chen2019uniter} as the representative model for single-stream, and LXMERT\footnote{\url{https://github.com/airsplay/lxmert}}  \cite{tan2019lxmert} for two-stream. As shown in Figure~\ref{fig:pretrained_model}(a), UNITER-base has the same model structure as the BERT-base model \cite{bert}, which composes of 12 layers of self-attention Transformers. Each layer has 12 self-attention heads, and each hidden representation is a 768-dimensional vector. 
As shown in Figure~\ref{fig:pretrained_model}(b),
LXMERT is a two-stream model that performs intra-attention in the same modality first, then cross-attention. We denote $T_t, T_v$, and $T_c$ as the Transformer modules that specifically model text-to-text, image-to-image and cross-modal interactions, respectively. In LXMERT, $T_t$ has 9 layers, $T_v$ has 5 layers, and $T_c$ has 5 layers. Each Transformer's hidden representation is in dimension of 768. Note that each layer in $T_c$ contains one cross-attention layer between two modalities, followed by two self-attention layers for each modality.

\section{Results on Untrained Baselines} \label{sec:untrained_results}

To measure the gain from learning, we also conducted additional experiments on untrained single-stream (SS) and two-stream (TS) baselines with random weights. 

\vspace{5pt}
\noindent \textbf{Multimodal Fusion Probe}\,
The untrained SS model has NMI of 0.99 for all output layers, suggesting that the two modalities are completely separated.
The untrained TS model has NMI of 0.56 for all output layers. This is because the cross-modality encoder layers
force the two modalities to fuse, even in untrained setting.

\vspace{5pt}
\noindent \textbf{Modality Importance Probe}\,
For the untrained model, the average attention of \texttt{[CLS]} token on the image/text modality is 0.66/0.28. Note that the
number of tokens in a sentence is usually smaller than that of the visual tokens.

\vspace{5pt}
\noindent \textbf{Visual Coreference and Relation Probe}\,
We provide additional untrained baselines for visual coreference and visual relation probes in Table~\ref{tab:untrained_baselines}.
Compared to Table~\ref{tab:visual_coreference} and Figure~\ref{fig:visual_relation_probing}(a), for VCD and VRI, untrained baselines for both SS and TS are equivalent to random
guess. For VRC, both SS and TS models outperform the baseline by around 10\%. For VCC, the SS model outperforms
the baseline by 17\%; while the TS model performs worse. This may be because after hard-designed multimodal
fusion, the direct coreference relationship between a pair of image/text tokens is diluted after training.

\begin{table}[t!]
\centering
\begin{tabular}{ccccc}
\toprule
\bf Model & \bf VCD & \bf VCC & \bf VRI & \bf VRC \\
\midrule
Untrained SS & 50.0  & 58.0 & 50.0 & 11.4 \\
Untrained TS & 50.0 & 66.0 & 50.0 & 9.34 \\
\bottomrule
\end{tabular} \\
\vspace{2pt}
\caption{\label{tab:untrained_baselines}\small{Untrained visual coreference/relation attention baselines.}}
\vspace{-3mm}
\end{table}

Furthermore, we provide an additional evaluation on whether the head selected for a specific coreference relation of an image-text pair
imposes higher attention scores for coreference relation than all other pairs. Results are summarized in Table~\ref{tab:additional_visual_coref_probe}, which suggests that these attention heads with maximum attention weight do pay more attention to the coreference image-text pair, compared to other unpaired ones.

\begin{table}[t!]
\centering
\begin{tabular}{ccccccc}
\toprule
\bf Coref Type & \tt people & \tt body parts & \tt scene & \tt clothing & \tt instruments & \tt animals \\
\midrule
Ratio & 0.33  & 0.23 & 0.28 & 0.37 & 0.59 & 0.53 \\
\bottomrule
\end{tabular} \\
\vspace{2pt}
\caption{\label{tab:additional_visual_coref_probe}\small{Results on whether the head selected for a specific coreference relation between an image-text pair
imposes higher attention scores than all the other pairs.}}
\vspace{-7mm}
\end{table}

\section{Additional Guidelines for Future Model Design} \label{sec:additional_guidelines}

In addition to the key takeaways in Sec.~\ref{sec:conclusion}, we provide a set of guidelines for future model design based on our analysis and observations.

\vspace{5pt}
($i$) Single-stream model is able to capture sufficient intra- and cross-modal knowledge, while the restricted attention
structure in two-stream model does not bring additional benefit. For future work, we will further explore single-stream model design,
which also exhibits better interpretability as observed.

($ii$) Initializing V+L model with BERT's weights should be helpful, which can enhance V+L model's capability in language understanding.

($iii$) It remains unclear how to measure a pre-trained model without evaluating on downstream tasks. Given that finetuning is time consuming,
the probing tasks we propose can provide a convenient tool to quickly test intermediate model
checkpoints during pre-training.

($iv$) Explicitly adding extra supervision to probing tasks during model training may lead to more interpretable and
robust model.

\section{Details on Linguistic Probe}
We probe the pre-trained models over nine tasks defined in the SentEval toolkit~\cite{conneau2018senteval}, under three categories: 
\vspace{5pt}
 
($i$) \emph{Surface tasks}: probe for the length of a sentence (\texttt{SentLen});

($ii$) \emph{Syntactic tasks}: predict the depth of a sentence's syntax tree, consecutive token inversions (\texttt{BShift}), and the top constituents sequences (\texttt{TopConst});
   
($iii$) \emph{Semantic tasks}: test the tense (\texttt{Tense}), the number implied by the subject/object (\texttt{SubjNum}/\texttt{ObjNum}), the replacement of the noun/verb form (\texttt{SOMO}), and the inversion of coordinating conjunctions (\texttt{CoordInv}).

\section{Additional Results} \label{sec:exp_results}
We provide additional results on multimodal fusion, visual coreference resolution, visual relation detection, and linguistic probing. 

\subsection{An t-SNE Visualization of Multimodal Fusion Degree}

\begin{figure}[h!]
    \centering
    \includegraphics[width=\linewidth]{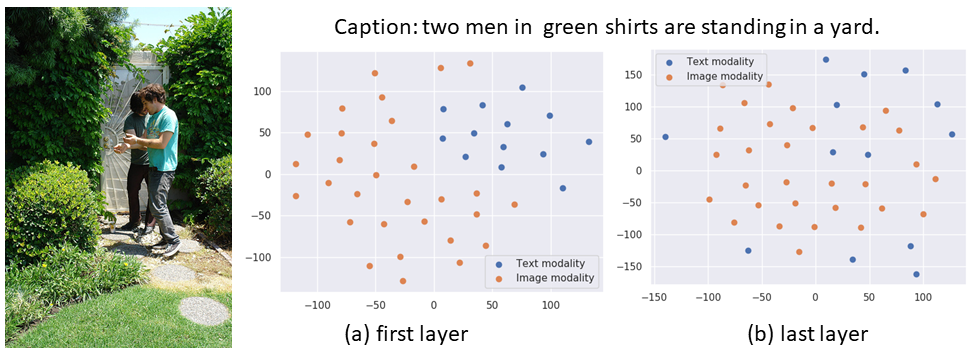}
    \caption{\label{fig:multimodal_fusion}\small{An t-SNE visualization of multimodal fusion degree of the first and last layer of UNITER over one image-text pair. Each yellow and blue dot corresponds to a visual and textual token, respectively.}}
\end{figure}

An t-SNE visualization of multimodal fusion degree of the first and last layer of UNITER (over one image-text pair) is provided in Figure~\ref{fig:multimodal_fusion}. As the layer goes deeper, the two modalities become more intertwined. 

\subsection{Visual Coreference Resolution}
Due to space limit, we only reported results using the embeddings from Layer 1, 5 and 12 in Table~\ref{tab:visual_coreference}. A complete set of results is provided in Table~\ref{tab:visual_coreference_more}. We observe that the attention probers work well for VCC, but not for VCD. Our assumption is that task granularity matters to the prober’s performance. Attention behavior varies a lot in different coreference relations, thus it
performs well on VCC. The dataset for training VCC is built with positive examples from VCD only. Therefore, VCD's settings naturally dilute the distinction between different coreference relations' attentions, which makes it a more challenging task.

\input{tables/vcr_results_complete}

\subsection{Visual Relation Detection}

Results of the layer-wise embedding probers on the Visual Relation Classification and Identification (VRC and VRI) tasks are visualized in Figure~\ref{fig:visual_relation_probing}(b) and (c), respectively. Detailed numbers corresponding to these two figures are provided in Table~\ref{tab:visual_relation_ss} and \ref{tab:visual_relation_ts}.

\input{tables/vrd_results_ss}
\input{tables/vrd_results_ts}

\subsection{Linguistic Probing} 
For linguistic probing, we first obtain results from all the layers of a pre-trained model, then report the best number in Table~\ref{tab:lingustic_probing}. Detailed results for all the layers are provided in Table~\ref{tab:lingustic_probing_more}.

\input{tables/linguistic_probe_complete}

\subsection{Visualization of Attention Maps}  \label{sec:visualization}

\begin{figure}[t!]
\centering
\begin{subfigure}[b]{4.25cm}            
\includegraphics[width=4.25cm]{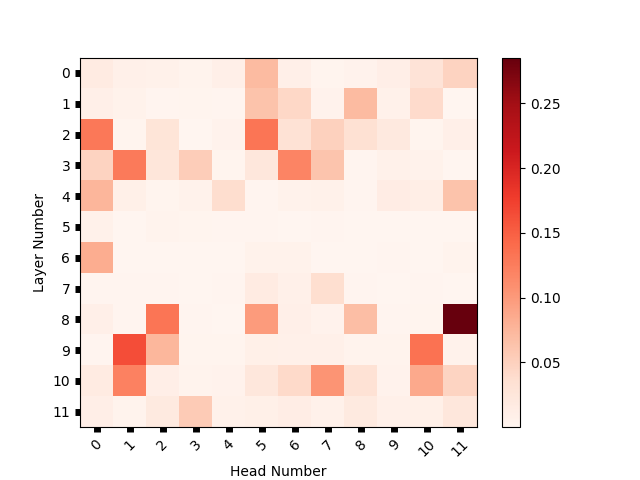}
\caption{\small{relation: \textit{animal}}}
\end{subfigure}
\hspace{0.5cm}
\begin{subfigure}[b]{4.25cm}
\centering
\includegraphics[width=4.25cm]{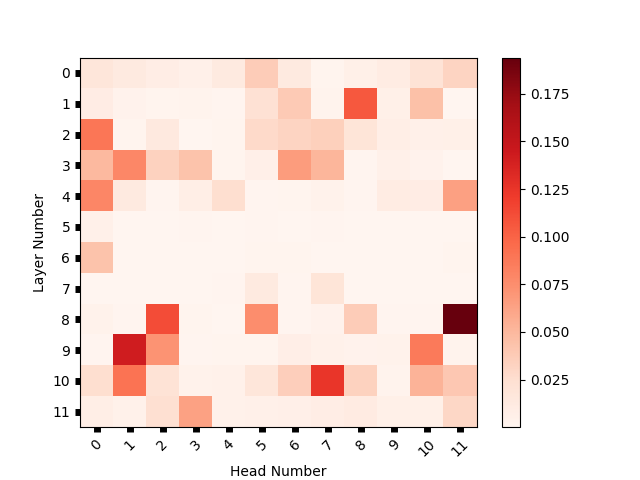}
\caption{\small{relation: \textit{vehicle}}}
\end{subfigure}
\vspace{0.5cm}
\begin{subfigure}[b]{4.25cm}
\centering
\includegraphics[width=4.25cm]{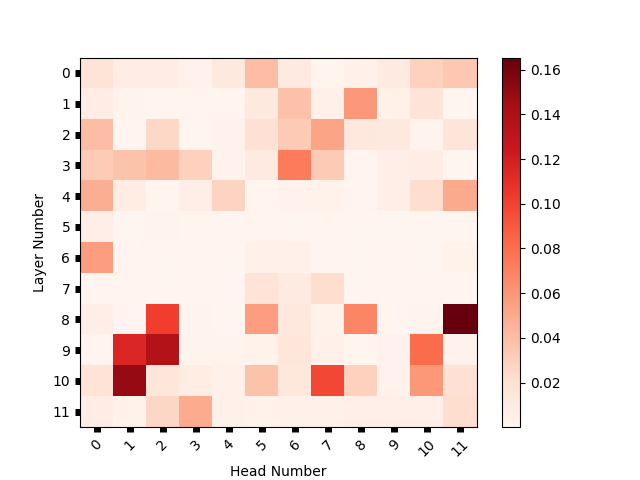}
\caption{\small{relation: \textit{people}}}
\end{subfigure}
\hspace{0.5cm}
\begin{subfigure}[b]{4.25cm}
\centering
\includegraphics[width=4.25cm]{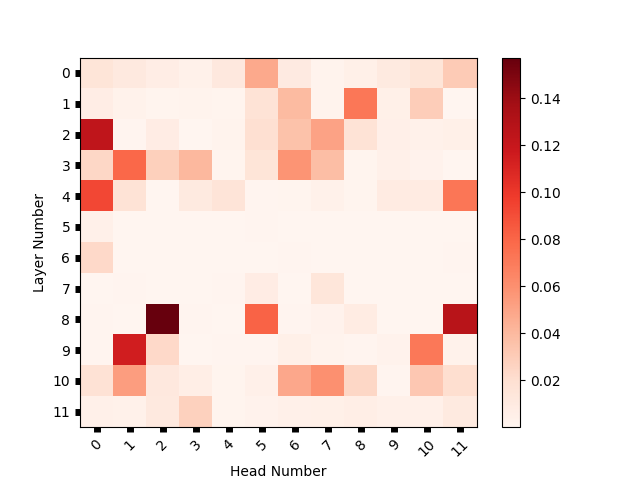}
\caption{\small{relation: \textit{clothing}}}
\end{subfigure}
\vspace{-3mm}
\caption{\label{fig:vis_coref_ss}\small{Visualization of coreference information for all 144 attention heads $(V \to T)$ in the single-stream model (UNITER-base). Note that only a set of attention heads is significant to the $V \to T$ attention across different coreference relations.}}
\end{figure}

\begin{figure}[t!]
\centering
\begin{subfigure}[b]{4.25cm}            
\includegraphics[width=4.5cm]{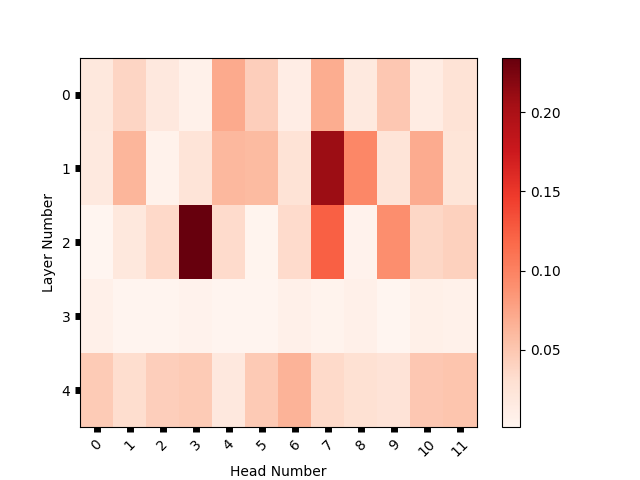}
\caption{\small{relation: \textit{animal}}}
\end{subfigure}
\hspace{0.5cm}
\begin{subfigure}[b]{4.25cm}
\centering
\includegraphics[width=4.25cm]{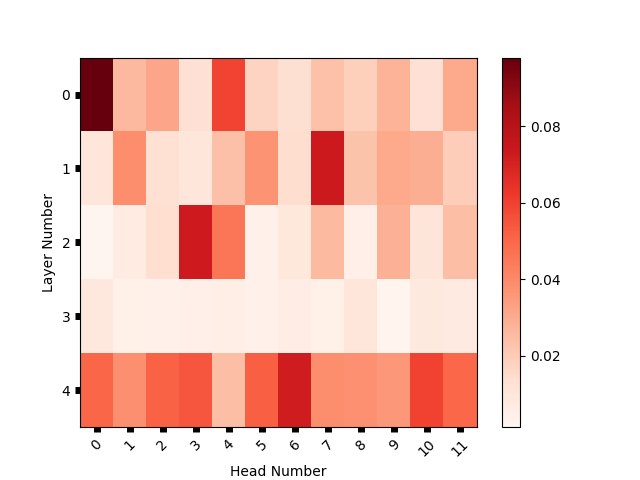}
\caption{\small{relation: \textit{vehicle}}}
\end{subfigure}
\vspace{0.5cm}
\begin{subfigure}[b]{4.25cm}
\centering
\includegraphics[width=4.25cm]{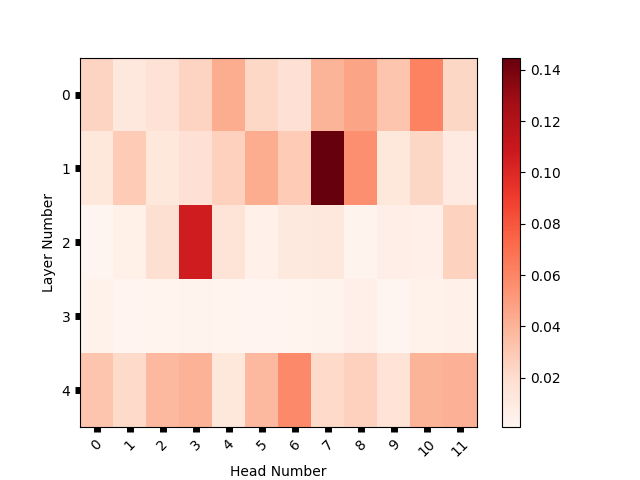}
\caption{\small{relation: \textit{people}}}
\end{subfigure}
\hspace{0.5cm}
\begin{subfigure}[b]{4.25cm}
\centering
\includegraphics[width=4.25cm]{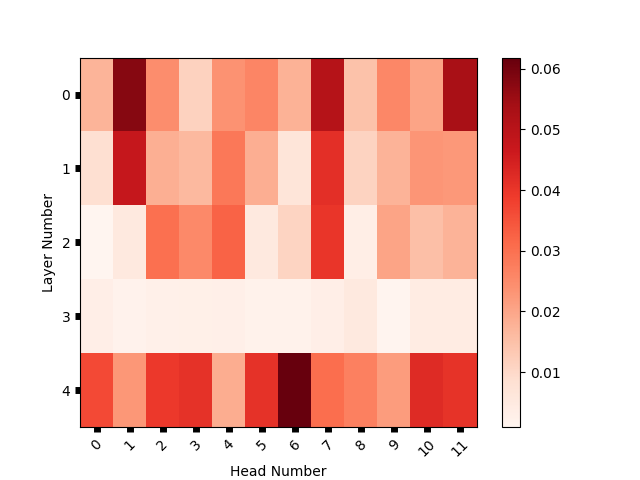}
\caption{\small{relation: \textit{clothing}}}
\end{subfigure}
\vspace{-3mm}
\caption{\label{fig:vis_coref_ts}\small{Visualization of coreference information across all attention heads $(V \to T)$ in the two-stream model (LXMERT-base, 5 layers, 12 heads per layer).}}
\end{figure}

\begin{figure}[t!]
\centering
\begin{subfigure}[b]{4.25cm}            
\includegraphics[width=4.25cm]{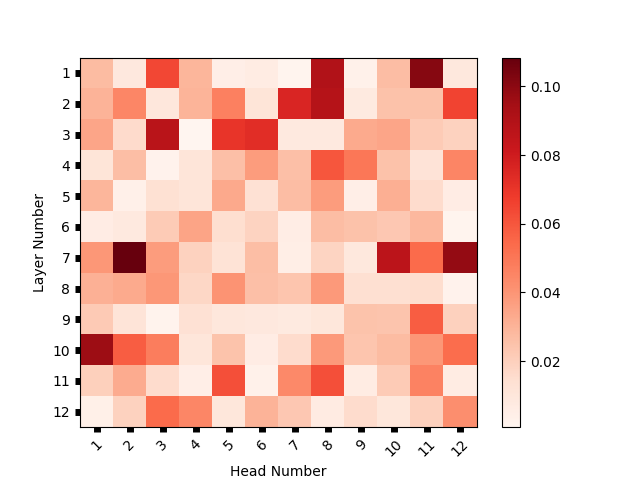}
\caption{\small{relation: \textit{covering}}}
\end{subfigure}
\hspace{0.5cm}
\begin{subfigure}[b]{4.25cm}
\centering
\includegraphics[width=4.25cm]{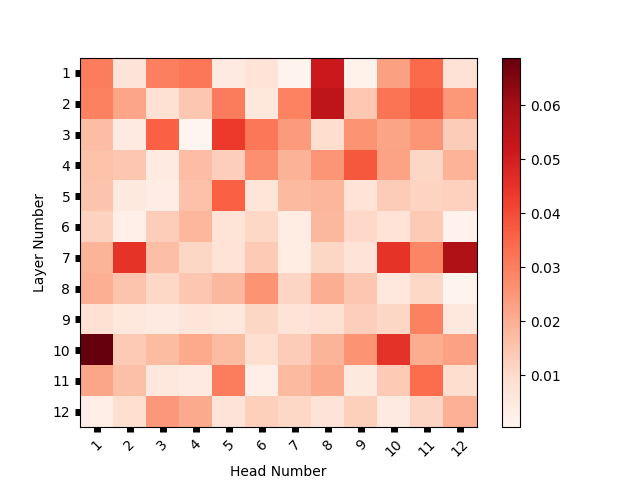}
\caption{\small{relation: \textit{at}}}
\end{subfigure}
\vspace{0.5cm}
\begin{subfigure}[b]{4.25cm}
\centering
\includegraphics[width=4.25cm]{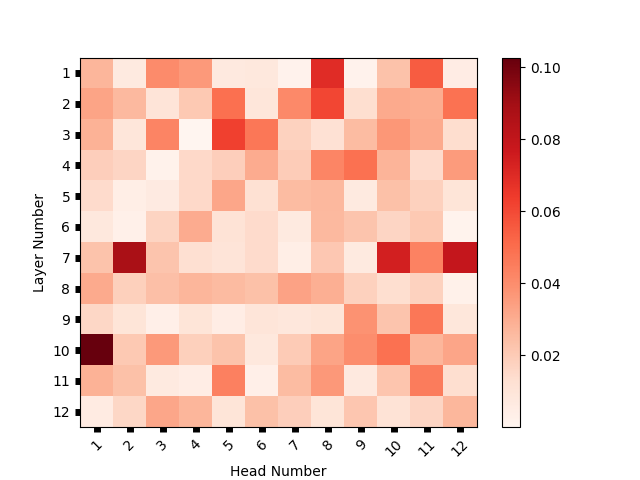}
\caption{\small{relation: \textit{on}}}
\end{subfigure}
\hspace{0.5cm}
\begin{subfigure}[b]{4.25cm}
\centering
\includegraphics[width=4.25cm]{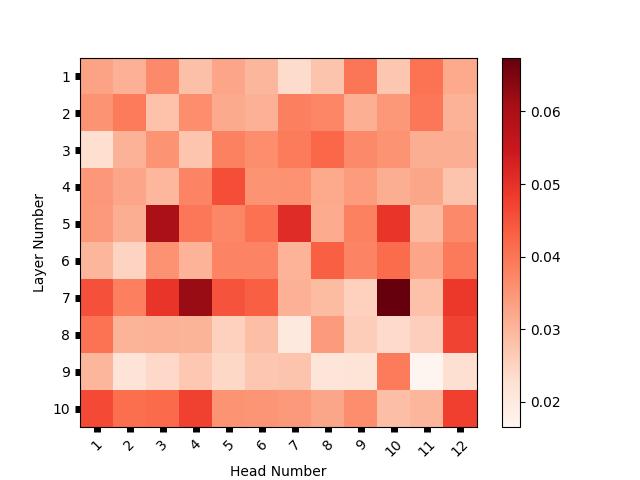}
\caption{\small{relation: \textit{playing}}}
\end{subfigure}
\vspace{-4mm}
\caption{\label{fig:vis_relation_ss}\small{Visualization of the maximum attention between two visually-related tokens across 144 attention heads in single-stream model (12 layers, 12 heads per layer). Note that the spatial relationships (\texttt{on}, \texttt{at}) have similar attention maps compared to other relations.}}
\end{figure}

\begin{figure}[t!]
\centering
\begin{subfigure}[b]{4.25cm}            
\includegraphics[width=4.25cm]{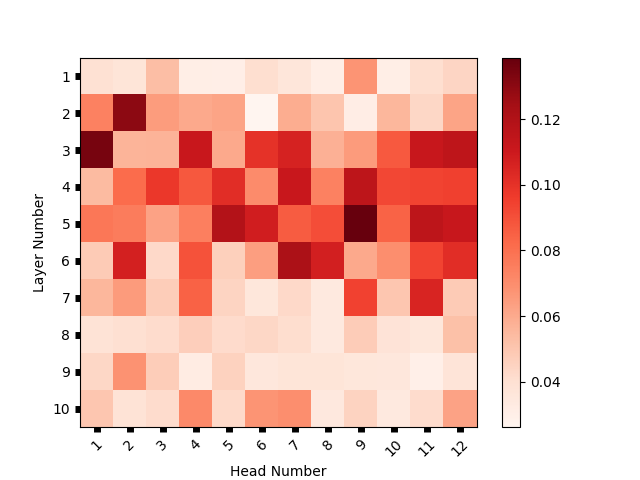}
\caption{\small{relation: \textit{covering}}}
\end{subfigure}
\hspace{0.5cm}
\begin{subfigure}[b]{4.25cm}
\centering
\includegraphics[width=4.25cm]{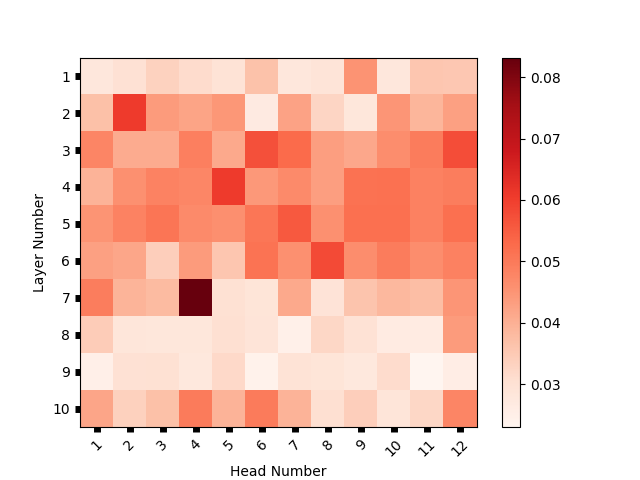}
\caption{\small{relation: \textit{at}}}
\end{subfigure}
\vspace{0.5cm}
\begin{subfigure}[b]{4.25cm}
\centering
\includegraphics[width=4.25cm]{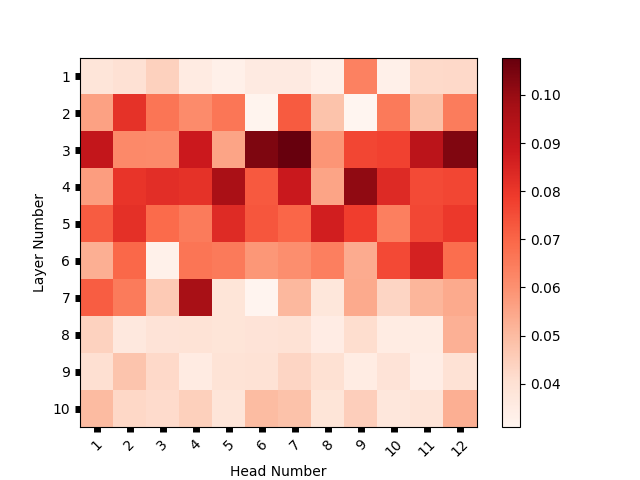}
\caption{\small{relation: \textit{on}}}
\end{subfigure}
\hspace{0.5cm}
\begin{subfigure}[b]{4.25cm}
\centering
\includegraphics[width=4.25cm]{figures/ts_playing_max_attention.png}
\caption{\small{relation: \textit{playing}}}
\end{subfigure}
\vspace{-4mm}
\caption{\label{fig:vis_relation_ts}\small{Visualization of the maximum attention between two visually-related tokens across the attention heads in two-stream model (10 layers: 1-5 layers: self-attention; 6-10 layers: cross-self-attention, 12 heads per layer).}}
\end{figure}

We show the learned attention maps of one specific relation in the probing tasks: Figure~\ref{fig:vis_coref_ss} and \ref{fig:vis_coref_ts} for visual coreference resolution (Section 3.3.2 of the main paper), and Figure~\ref{fig:vis_relation_ss} and \ref{fig:vis_relation_ts} for visual relation detection (Section 3.4 of the main paper). 

%% file: tables/vcr_results_complete.tex
\begin{table}[t!]
    \centering
\begin{tabular}{ccccc}
\toprule
\bf Classifier Input & \bf VCD (SS) & \bf VCD (TS) & \bf VCC (SS) & \bf VCC (TS)  \\
\midrule
144 Attention Heads & 52.04  & \textbf{53.68} & 75.10 & 54.47 \\
Random Guess & 50.00 & 50.00 & 12.50 & 12.50 \\
\midrule
Layer 1 & 56.86 & \textbf{53.68} & 93.51 & 93.35  \\
Layer 2 & 57.58 & 53.49 & 93.91 & \textbf{93.36}  \\
Layer 3 & 57.81 & 53.32 & 94.11 & 93.32  \\
Layer 4 & 57.97 & 52.92 & 94.10 & 93.12  \\
Layer 5 & \textbf{59.12} & 52.59 & 94.05 & 92.62 \\
Layer 6 & 58.58 & / & 94.02 & /  \\
Layer 7 & 58.67 & / & \textbf{94.26} & / \\
Layer 8 & 58.65 & / & 93.96 & / \\
Layer 9 & 58.15 & / & 93.77 & / \\
Layer 10 & 57.65 & / & 93.77 & / \\
Layer 11 & 57.96 & / & 93.47 & / \\
Layer 12 & 58.40 & / & 93.44 & / \\
\bottomrule
\end{tabular} \\
\caption{\label{tab:visual_coreference_more}\small{Results of attention and layer-wise embedding probers on Visual Coreference Detection and Classification (VCD and VCC). SS: single-stream; TS: two-stream.}}
\end{table}

%% file: tables/vrd_results_ss.tex
\begin{table}[t!]
\centering
\begin{tabular}{ccccccc}
\toprule
\bf Classifier Input & \bf VRI (SS) & \bf VRC (SS) & \bf VRI (SS mis.) & \bf VRC (SS mis.) \\
\midrule
Original visual emb. & 76.95 & 36.38 & 76.95 & 36.38 &   \\
Layer 1 & 77.18 & 37.70 & \bf 77.03 & 37.31 \\
Layer 2 & 79.56 & 42.22 & 76.84 & 37.55 \\
Layer 3 & 82.28 & 43.02 & 76.55 & 37.05 \\
Layer 4 & 83.24 & 45.88 & 76.40 & 37.66 \\
Layer 5 & 84.45 & 47.67 & 76.28 & 37.49 \\
Layer 6 & 84.35 & 46.46 & 75.99 & 37.54 \\
Layer 7 & 84.13 & 45.67 & 75.88 & 37.51 \\
Layer 8 & 83.95 & 45.32 & 75.62 & \bf 37.71 \\
Layer 9 & 85.98 & 54.35 & 74.75 & \bf 37.71\\
Layer 10 & \bf 86.35 & 55.66 & 74.15 & 37.06 \\
Layer 11 & 86.19 & \bf 56.64 & 73.84 & 36.72 \\
Layer 12 & 86.07 & 55.22 & 72.96 & 36.65\\
\bottomrule
\end{tabular}
\caption{\small{\label{tab:visual_relation_ss}Accuracies (\%) of the layer-wise embedding probers on Visual Relation Identification and Classification (VRI and VRC) tasks for the single-stream (SS) model. mis.: mismatch.}}
\end{table}

%% file: tables/vrd_results_ts.tex
\begin{table}[t!]
\centering
\begin{tabular}{ccccc}
\toprule
\bf Classifier Input & \bf VRI (TS) & \bf VRC (TS) & \bf VRI (TS mis.) & \bf VRC (TS mis.) \\
\midrule
Original visual emb. & 76.95 & 36.38 & 76.95 & 36.38  \\
Layer 1 & 75.92 & 36.61 & 77.58 & \bf 36.57 \\
Layer 2 & 75.42 & 35.86 & 77.44 & 35.82 \\
Layer 3 & 75.01 & 35.72 & 77.38 & 35.66 \\
Layer 4 & 74.67 & 36.01 & 77.19 & 35.99 \\
Layer 5 & 74.75 & 36.45 & 77.05 & 36.43 \\
Layer 6 & \bf 87.00 & 67.82 & \bf 87.03 & 13.20 \\
Layer 7 & 86.59 & 68.06 & 86.24 & 12.83 \\
Layer 8 & 86.14 & 68.67 & 85.86 & 11.93 \\
Layer 9 & 85.50 & 68.57 & 85.36 & 12.11 \\
Layer 10 & 85.43 & \bf 69.66 & 85.26 & 12.07 \\
\bottomrule
\end{tabular}
\caption{\small{\label{tab:visual_relation_ts}Accuracies (\%) of the layer-wise embedding probers on Visual Relation Identification and Classification (VRI and VRC) tasks for the two-stream (TS) model. mis.: mismatch.}}
\end{table}

%% file: tables/linguistic_probe_complete.tex
\begin{table}[t!]
\centering
\small
\resizebox{0.95\textwidth}{!}{%
\begin{tabular}{ c*{5}{|c}}
\toprule
\multicolumn{1}{l} {\bf Layer}  & \multicolumn{1}{l} {\bf SentLen} & \multicolumn{1}{l} {\bf TreeDepth} & \multicolumn{1}{l} {\bf TopConst} & \multicolumn{1}{l} {\bf BShift} & \multicolumn{1}{l} {\bf Tense}  \\
 \multicolumn{1}{l} {} &   \multicolumn{1}{l} {(Surface)} & 
\multicolumn{1}{l} {(Syntactic)} &  \multicolumn{1}{l} {(Syntactic)} &  \multicolumn{1}{l} {(Syntactic)} &  \multicolumn{1}{l} {(Semantic)} \\
\midrule
1 & 86.5, 75.8, \bf 93.9 & 29.8, 29.3, \bf 35.9 &  36.5, 31.8, \bf 63.6 & 50.0, 50.0, \bf 50.3 & 71.8, 66.0, \bf 82.2  \\
2 & 88.8, 73.8, \bf 95.9 & 33.6, 27.9, \bf 40.6 &  54.6, 29.4, \bf 71.3 & 50.0, 50.0, \bf 55.8 & 77.8, 70.5, \bf 85.9  \\
3 & 87.4, 74.4, \bf 96.2 & 34.7, 28.0, \bf 39.7 &  67.5, 32.5, \bf 71.5 & 56.6, 51.8, \bf 64.9 & 83.7, 72.1, \bf 86.6 \\
4 & 87.7, 76.6, \bf 94.2 & 35.3, 29.0, \bf 39.4 &  69.7, 39.3, \bf 71.3 & 71.2, 54.9, \bf 74.4 & 84.3, 72.5, \bf 87.6 \\
5 & 86.5, 77.5, \bf 92.0 & 36.4, 29.5,  \bf 40.6 &  72.5, 48.6, \bf 81.3 & 73.6, 55.4, \bf 81.4 & 84.1, 74.4, \bf 89.5 \\
6 & 85.0, 81.8, \bf 88.4 & 36.1, 31.5, \bf 41.3 &  73.5, 48.1, \bf 83.3 & 74.6, 62.3, \bf 82.9 & 83.0, 75.6, \bf 89.8 \\
7 & 83.6,  \textbf{83.8}, 83.7 & 36.2, 32.7, \bf 40.1 & 79.0, 63.4, \bf 84.1 & 76.5, 63.4, \bf 83.0 & 83.8, 75.6, \bf 89.9 \\
8 & 81.7, 81.8, \bf 82.9 & 35.1, 34.0, \bf 39.2 & 78.0, 67.2, \bf 84.0  & 77.3, 64.9, \bf 83.9 & 84.0, 75.2, \bf 89.9 \\
9 & 79.7, 79.8, \bf 80.1 & 34.5, 32.7, \bf 38.5 &  76.5, 65.7, \bf 83.1 & 78.8, 64.8, \bf 87.0 & 85.3, 75.1, \bf 90.0 \\
10 & \textbf{77.4}, \hspace{0.7mm} / \hspace{0.7mm}, 77.0 & 33.9, \hspace{0.7mm} / \hspace{0.7mm}, \bf 38.1 & 75.6, \hspace{0.7mm} / \hspace{0.7mm}, \bf 81.7 & 81.6, \hspace{0.7mm} / \hspace{0.7mm}, \bf 86.7 & 86.4, \hspace{0.7mm} / \hspace{0.7mm}, \bf 89.7 \\
11 & \textbf{77.5}, \hspace{0.7mm} / \hspace{0.7mm}, 73.9 & 34.1, \hspace{0.7mm} / \hspace{0.7mm}, \bf 36.3 & 73.9, \hspace{0.7mm} / \hspace{0.7mm}, \bf 80.3 & 80.9, \hspace{0.7mm} / \hspace{0.7mm}, \bf 86.8 & 86.6, \hspace{0.7mm} / \hspace{0.7mm}, \bf 89.9 \\
12 & \textbf{74.6}, \hspace{0.7mm} / \hspace{0.7mm}, 69.5 & 32.2, \hspace{0.7mm} / \hspace{0.7mm}, \bf 34.7 & 70.9, \hspace{0.7mm} / \hspace{0.7mm}, \bf 76.5 & 80.8, \hspace{0.7mm} / \hspace{0.7mm}, \bf 86.4 & 86.2, \hspace{0.7mm} / \hspace{0.7mm}, \bf 89.5  \\
\midrule
\toprule
\multicolumn{1}{l} {\bf Layer}  & \multicolumn{1}{l} {\bf SubjNum}  & \multicolumn{1}{l} {\bf ObjNum}  & \multicolumn{1}{l} {\bf SOMO}  & \multicolumn{1}{l} {\bf CoordInv} \\
 \multicolumn{1}{l} {} & \multicolumn{1}{l} {(Semantic)} &  \multicolumn{1}{l} {(Semantic)} &  \multicolumn{1}{l} {(Semantic)} &  \multicolumn{1}{l} {(Semantic)} \\
\midrule
1 & 69.0, 70.6, \bf 77.6 & 65.3, 69.1, \bf 76.7 & 49.9, \textbf{51.0}, 49.9 & 50.0, 51.2, \bf 53.9 \\
2 & 74.8, 71.2, \bf 82.5 & 72.5, 70.8, \bf 80.6 & 50.1, 50.0, \bf 53.8 & 50.5, 50.0, \bf 58.5 \\
3 & 79.6, 70.7, \bf 82.0 & 78.2, 70.0, \bf 80.3 & 50.3, 50.5, \bf 55.8 & 56.8, 50.1, \bf 59.3\\
4 & 79.9, 72.6, \bf 81.9 & 77.0, 71.7, \bf 81.4 & 50.9, 50.1, \bf 59.0 & 57.8, 50.0, \bf 58.1\\
5 & 80.5, 74.8, \bf 85.8 & 76.6, 74.5, \bf 81.2 & 51.0, 50.1, \bf 60.2 & 59.4, 56.2, \bf 64.1\\
6  & 81.1, 74.6, \bf 88.1 & 77.1, 74.6, \bf 82.0 & 52.2, 50.3, \bf 60.7 & 59.4, 56.1, \bf 71.1\\
7 & 83.4, 77.3, \bf 87.4 & 78.4, 76.0, \bf 82.2 & 54.3, 51.2, \bf 61.6 & 60.3, 57.4, \bf 74.8\\
8  & 82.7, 78.8, \bf 87.5 & 78.2, 76.8, \bf 81.2 & 54.5, 51.4, \bf 62.1 & 60.5, 58.7, \bf 76.4\\
9  & 81.8, 78.8, \bf 87.6 & 78.8, 76.7, \bf 81.8 & 55.9, 51.0, \bf 63.4 & 61.8, 58.0, \bf 78.7\\
10 & 81.9, \hspace{0.7mm} / \hspace{0.7mm}, \bf 87.1 & 78.5, \hspace{0.7mm} / \hspace{0.7mm}, \bf 80.5 & 56.3, \hspace{0.7mm} / \hspace{0.7mm}, \bf 63.3 & 61.9, \hspace{0.7mm} / \hspace{0.7mm}, \bf 78.4\\
11 & 83.0, \hspace{0.7mm} / \hspace{0.7mm}, \bf 85.7 & 78.2, \hspace{0.7mm} / \hspace{0.7mm}, \bf 78.9 & 56.6, \hspace{0.7mm} / \hspace{0.7mm}, \bf 64.4  & 62.1, \hspace{0.7mm} / \hspace{0.7mm}, \bf 77.6\\
12 & 81.8, \hspace{0.7mm} / \hspace{0.7mm}, \bf 84.0 & 77.8, \hspace{0.7mm} / \hspace{0.7mm}, \bf 78.7 & 57.1, \hspace{0.7mm} / \hspace{0.7mm}, \bf 65.2 & 61.7, \hspace{0.7mm} / \hspace{0.7mm}, \bf 74.9 \\
\Xhline{2\arrayrulewidth} 
\end{tabular}}
\caption{\label{tab:lingustic_probing_more}\small{Results on the linguistic probing tasks. For each task and each layer, the results are presented in the order of UNITER, LXMERT, and the original BERT.}}
\end{table} 